\documentclass[letterpaper]{article} 
\usepackage{aaai23}  
\usepackage{times}  
\usepackage{helvet}  
\usepackage{courier}  
\usepackage[hyphens]{url}  
\usepackage{graphicx} 
\graphicspath{{images_cmyk_U/}} 
\usepackage{natbib}  
\usepackage{caption} 
\usepackage{algorithm}
\usepackage{algorithmic}

\usepackage{newfloat}
\usepackage{listings}
\DeclareCaptionStyle{ruled}{labelfont=normalfont,labelsep=colon,strut=off} 
\floatstyle{ruled}
\newfloat{listing}{tb}{lst}{}
\floatname{listing}{Listing}
\usepackage{amssymb, amsmath}  
\usepackage{booktabs, multirow} 
\title{ParaFormer: Parallel Attention Transformer for Efficient Feature Matching}
\author {
    Xiaoyong Lu\textsuperscript{\rm 1},
    Yaping Yan\textsuperscript{\rm 1}, 
    Bin Kang\textsuperscript{\rm 2}, 
	Songlin Du\textsuperscript{\rm 1}\thanks{Corresponding author}
}
\affiliations {
    \textsuperscript{\rm 1}Southeast University, Nanjing, China\\
    \textsuperscript{\rm 2}Nanjing University of Posts and Telecommunication, Nanjing, China\\
	\{220211846, yan, sdu\}@seu.edu.cn, kangb@njupt.edu.cn
}

\usepackage{bibentry}

\begin{document}

\maketitle

\begin{abstract}

Heavy computation is a bottleneck limiting deep-learning-based feature matching algorithms to be applied in many real-time applications.
However, existing lightweight networks optimized for Euclidean data cannot address classical feature matching tasks, since sparse keypoint based descriptors are expected to be matched.
This paper tackles this problem and proposes two concepts: 1) a novel parallel attention model entitled ParaFormer and 2) a graph based U-Net architecture with attentional pooling.
First, ParaFormer fuses features and keypoint positions through the concept of amplitude and phase, and integrates self- and cross-attention in a parallel manner which achieves a win-win performance in terms of accuracy and efficiency.
Second, with U-Net architecture and proposed attentional pooling, the ParaFormer-U variant significantly reduces computational complexity, and minimize performance loss caused by downsampling.
Sufficient experiments on various applications, including homography estimation, pose estimation, and image matching, demonstrate that ParaFormer achieves state-of-the-art performance while maintaining high efficiency.
The efficient ParaFormer-U variant achieves comparable performance with less than 50\% FLOPs of the existing attention-based models. 

\end{abstract}

\section{Introduction}
\label{Introduction}
Feature matching is a fundamental problem for many computer vision tasks, such as object recognition \cite{sift-flow}, structure from motion (SFM) \cite{sfm}, and simultaneous localization and mapping (SLAM) \cite{slam}.
But with illumination changes, viewpoint changes, motion blur and occlusion, it is challenging to find the invariance and get robust matches from two images.

Feature matching pipelines can be categorized into detector-based methods, which first detect keypoints and descriptors from the images and then match two sets of sparse features, and detector-free methods, which directly match dense features.
Benefiting from the global modeling capability of Transformer \cite{transformer},
attention-based networks become dominant methods in both detector-based and detector-free pipelines, where self- and cross-attention are applied to match learning-based descriptors or dense features. 
However, despite the high performance, attention-based networks tend to bring high training costs, large memory requirements, and high inference latency,
especially for detector-free pipelines, where processing dense features exacerbates the problem of quadratic complexity of attention mechanism. 
So we focus on \emph{detector-based} pipeline, seeking the best trade-off between efficiency and performance. 

As most lightweight operations \cite{xception,mbnet} are designed for Euclidean data, sparse descriptors cannot be handled by mainstream lightweight networks.
Note that Transformer and Graph Neural Networks are suitable for processing non-Euclidean data,
so we design efficient models from both perspectives, giving birth to the ParaFormer and its ParaFormer-U variant.

\begin{figure}[t]
	\centering
	\includegraphics[width=0.99\columnwidth]{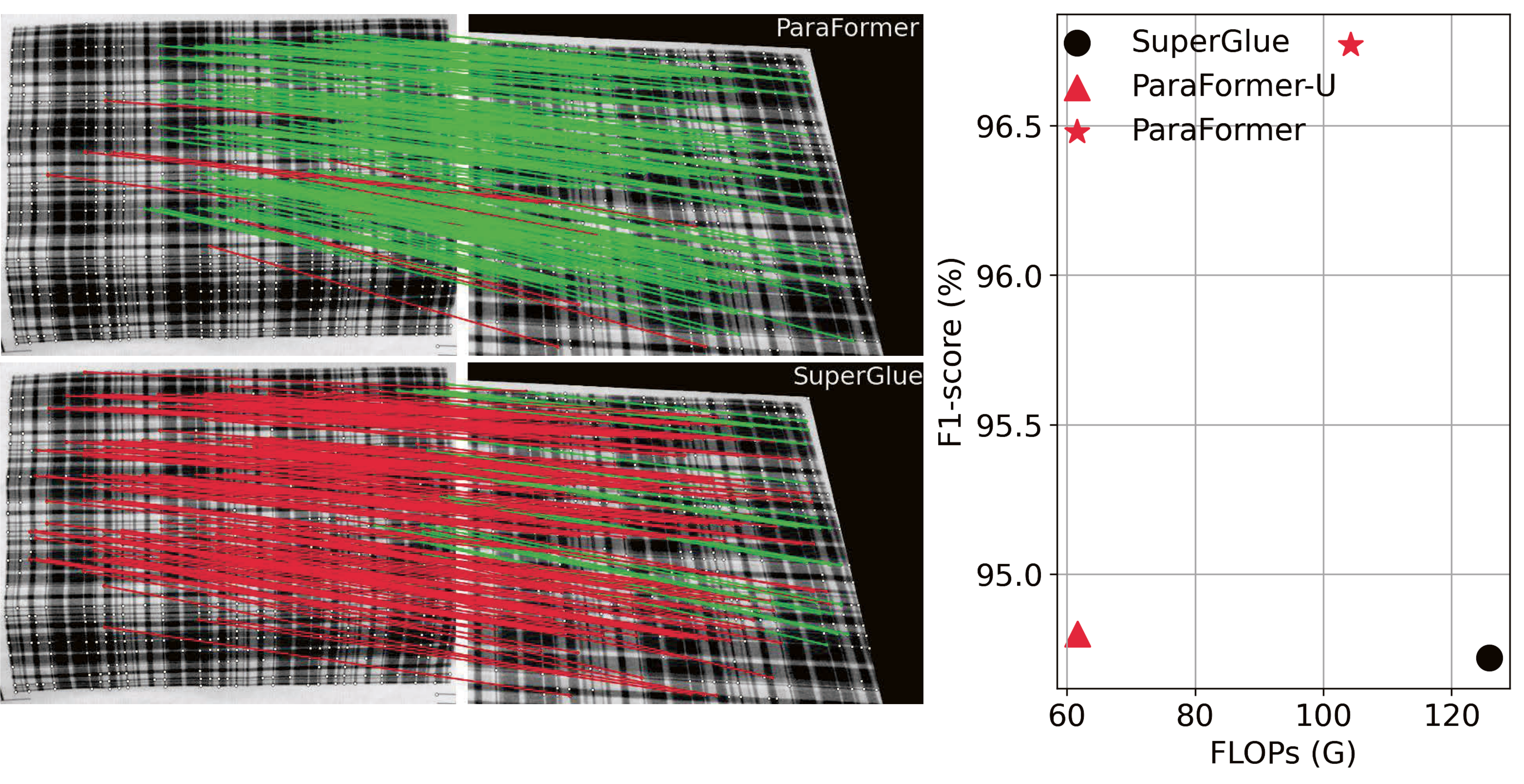}
	\caption{
	Comparison between the ParaFormer and SuperGlue.
	With the same input features, ParaFormer can deliver more robust matches with higher matching precision.
	ParaFormer-U can achieve comparable performance to SuperGLue with significantly fewer FLOPs.}
	\label{figure1}
\end{figure}

\begin{figure}[t]
	\centering
	\includegraphics[width=0.83\columnwidth]{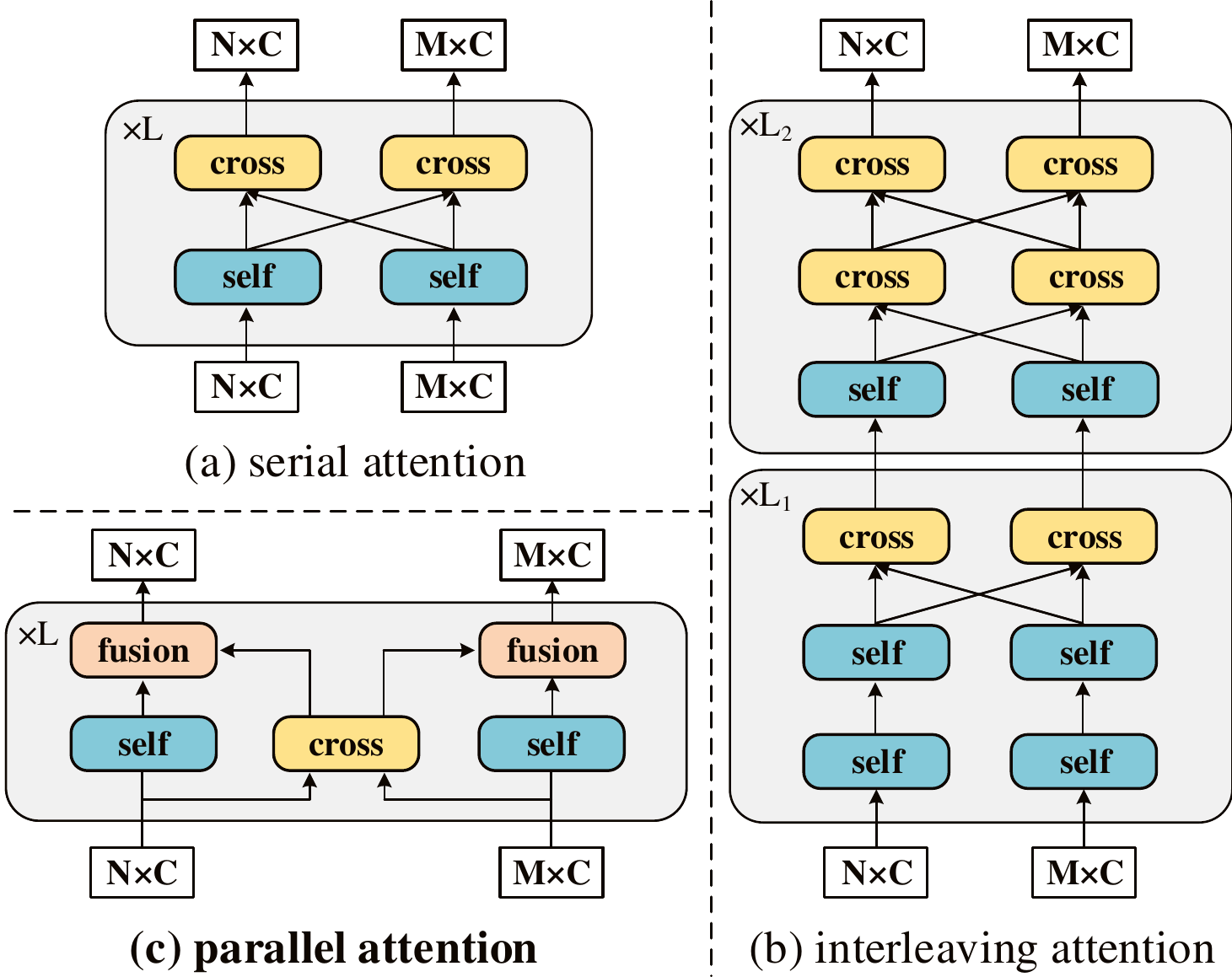}
	\caption{Conceptual difference among three attention architectures.
	(a) Serial attention in SuperGlue.
	(b) Interleaving attention in MatchFormer.
	(c) Proposed parallel attention.}
	\label{figure2}
\end{figure}

Rethinking the self- and cross-attention in feature matching,
all existing attention-based methods arrange two kinds of attention in a serial manner, a strategy derived from the behavior of people looking back and forth when matching images. 
Specifically, SuperGlue \cite{9157489} and LoFTR \cite{loftr} alternately arrange self- and cross-attention, \emph{i.e.}, the $self\rightarrow cross$ strategy as illustrated in Figure \ref{figure2} (a).
For MatchFormer \cite{matchformer}, the $self\rightarrow self\rightarrow cross$ strategy is used in the early stages, and the $self\rightarrow cross\rightarrow cross$ strategy is used in later stages as shown in Figure \ref{figure2} (b).
However, computer vision is not necessarily designed based on human behavior,
the fixed serial attention structure limits the diversity of the integration of self- and cross-attention.
We propose parallel attention to compute self- and cross-attention synchronously,
and train the network to optimally fuse two kinds of attention instead of tuning the permutation of both as a hyperparameter.

For the efficiency of the attention-based feature matching, instead of simply applying attention variants \cite{Linear, PVT},
weight sharing and attention weight sharing strategies in the parallel attention layer are explored to reduce redundant parameters and computations.
We further construct the U-Net architecture with parallel attention layers and propose attentional pooling, which identifies important context points by attention weights.

In summary, the contributions of this paper include:

\begin{itemize}
    \item We rethink the attention-based feature matching networks, and propose the \textbf{parallel attention layer} to perform self- and cross-attention synchronously and adaptively integrate both with learnable networks.
    \item We further explore the \textbf{U-Net architecture} for efficient feature matching and propose attentional pooling, which keeps only the important context points to reduce the FLOPs with minimal performance loss.
    \item A novel \textbf{wave-based position encoder} is proposed for detector-based feature matching networks, which dynamically fuses descriptors and positions through the concepts of amplitude and phase of waves.
\end{itemize}

\section{Related Works}
\label{Related Works}

\textbf{Local Feature Matching.}
Classical feature matching tends to be a detector-based pipeline, \emph{i.e.}, the detector is first applied to generate keypoints and descriptors from images, and then the descriptors are matched.
For detectors, some outstanding handcrafted methods \cite{2004Distinctive,surf,brief,orb} were first proposed and widely used for various 3D computer vision tasks. 
With the advent of the deep learning era, many learning-based detectors \cite{r2d2,superpoint,d2net,lfnet} have been proposed to further improve the robustness of descriptors under illumination changes and viewpoint changes.
In addition to detectors, other work has focused on better matchers.
SuperGlue \cite{9157489} was the first to propose an attention-based feature matching network that uses self- and cross-attention to find matches with global context information.
OETR \cite{oetr} further constrains attention-based feature matching in the commonly visible region by overlap estimation.

Besides matching the sparse descriptors generated by the detector, LoFTR \cite{loftr} applies self- and cross-attention directly on the feature maps extracted by convolutional neural network (CNN) and generates matches in a coarse-to-fine manner.
MatchFormer \cite{matchformer} further abandons the CNN backbone and adopts a completely attention-based hierarchical framework that can extract features while finding similarities utilizing the attention mechanism.
Noting that the permutation of self- and cross-attention in SuperGlue and LoFTR is a simple alternating strategy, MatchFormer further proposes an interleaving strategy,
which focuses on self-attention at the shallow stages of the network and cross-attention at the deep stages.
This improvement gives us inspiration about the permutation of self- and cross-attention.

All existing attention-based approaches artificially arrange self- and cross-attention in a serial manner to mimic human behavior, which does not take advantage of the benefits of deep learning network and parallel computing.
We propose to compute two kinds of attention efficiently in a parallel manner, and let the network learn the optimal way to integrate the two kinds of attention.

\begin{figure*}[t]
    \centering
    \includegraphics[width=0.9\textwidth]{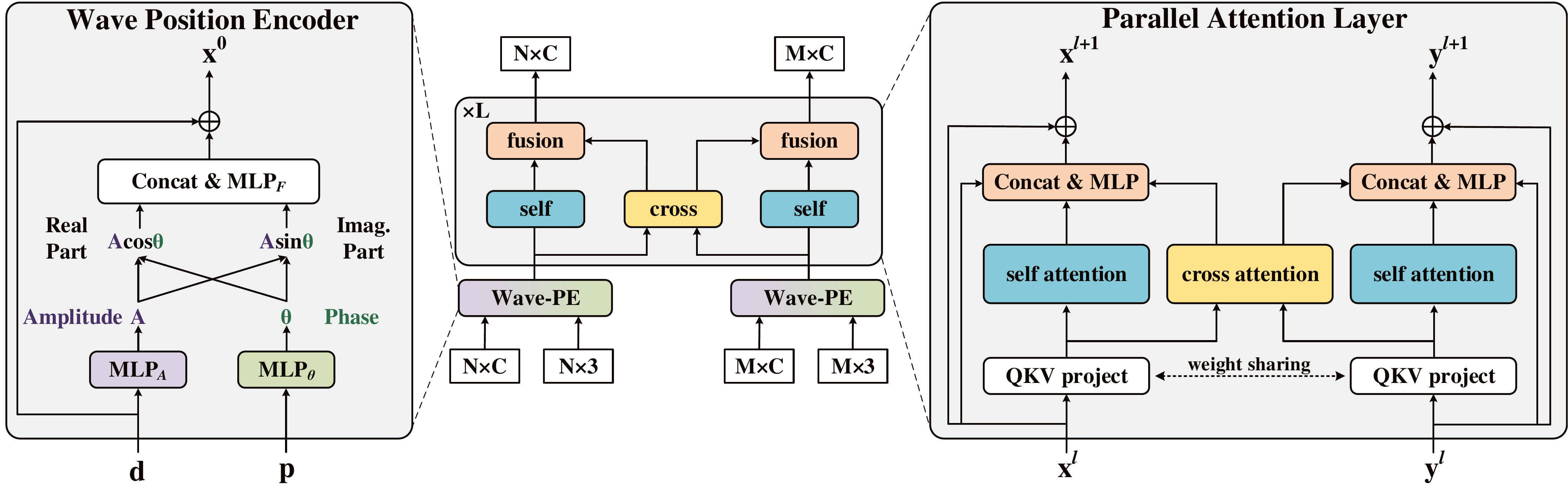} 
    \caption{ParaFormer architecture.
	Wave-PE fuses the amplitude $\mathbf{A}$ estimated with the descriptor $\mathbf{d}$ and the phase $\boldsymbol{\theta}$ estimated with the position $\mathbf{p}$ to generate position encoding.
	Stacked parallel attention layers utilize self- and cross-attention to enhance the descriptors and find potential matches, where self- and cross-attention are adaptively integrated through a learnable network.}
    \label{figure3}
\end{figure*}

\noindent{\textbf{Position Encoder.}}
The position encoder is a critical part for all transformer-based networks, which allows the network to sense the relative or absolute position of each vector in a sentence or image.
The first proposed position encoding method \cite{transformer} uses fixed sine and cosine functions to calculate the position encoding or uses the position encoding vector as a learnable parameter, and finally adds the position encoding to the original vector.
Although position information can be provided, this approach severely limits the flexibility of the model because the position encodings are fixed-length at training time,
which limits the model to only process fixed-length inputs at inference time.

Another way of position encoding is relative position encoding \cite{swin}, \emph{i.e.}, adjusting attention weights with relative position.
However, it is not only computationally intensive but also needs to handle inputs of different lengths by interpolation, which severely damages the performance.
Convolution-based position encoders \cite{cpe,matchformer} are proposed to augment local features with convolution and enable the model to be aware of position information with zero padding.
But this method can only be applied to Euclidean data such as feature maps, thus it cannot be applied to methods based on sparse descriptors.

To handle arbitrary length and non-Euclidean inputs, SuperGlue proposes a position encoder based on the multi-layer perceptron (MLP),
which uses MLP to extend the coordinate vector to align with the dimension of the descriptor to get the position encoding.
However, the weak encoding ability becomes the bottleneck of the matching network.
Inspired by Wave-MLP \cite{wave_mlp}, phase information is equally important in vectors compared to amplitude information. Wave-MLP encodes the same input as both amplitude and phase information,
while we encode the descriptor as amplitude information and the position as phase information, then fuse the two types of information with the Euler formula to generate position-aware descriptors.

\noindent{\textbf{U-Net Architecture.}}
The U-Net \cite{u_net} architecture consists of an encoder-decoder structure, where the encoder reduces the spatial resolution and the decoder recovers it.
This architecture can efficiently handle dense prediction tasks such as semantic segmentation, so we seek to improve the efficiency of attention-based feature matching with U-Net.
However conventional pooling operations cannot be directly applied to non-Euclidean data like sparse descriptors, so Graph U-Nets \cite{gu_net} proposes the graph pooling (gPool) layer to enable downsampling on graph data in a differentiable way.
The gPool layer measures the information that can be retained by each feature vector through scalar projection and applies topk sampling so that the new graph preserves as much information as possible from the original graph.
Based on the gPool layer, we propose to utilize attention weights to measure how much information can be retained by each feature vector, which can better cooperate with the attention-based network without introducing additional parameters.

\section{Methodology}
\label{Methodology}
Assuming that \emph{M} and \emph{N} keypoints are detected in image \emph{X} and image \emph{Y},
we let the positions be $\textbf{p}^X\in\mathbb{R}^{M\times 3}$, $\textbf{p}^Y\in\mathbb{R}^{N\times 3}$
and the descriptors be $\textbf{d}^X\in\mathbb{R}^{M\times C}$, $\textbf{d}^Y\in\mathbb{R}^{N\times C}$.
As illustrated in Figure \ref{figure3}, our proposed method first dynamically fuses positions $\textbf{p}$ and descriptors $\textbf{d}$ in amplitude and phase manner with wave position encoder (Wave-PE).
The parallel attention module is then applied to compute self- and cross-attention synchronously, utilizing global information to enhance the representation capability of features and find potential matches.
$\textbf{x}^{l}$ and $\textbf{y}^{l}$ denote the intermediate features of image \emph{X} and \emph{Y} in layer \emph{l}.
Finally, the enhanced descriptors are matched by the optimal matching layer \cite{9157489} appliying the Sinkhorn algorithm.

\begin{figure*}[t]
    \centering
    \includegraphics[width=0.85\textwidth]{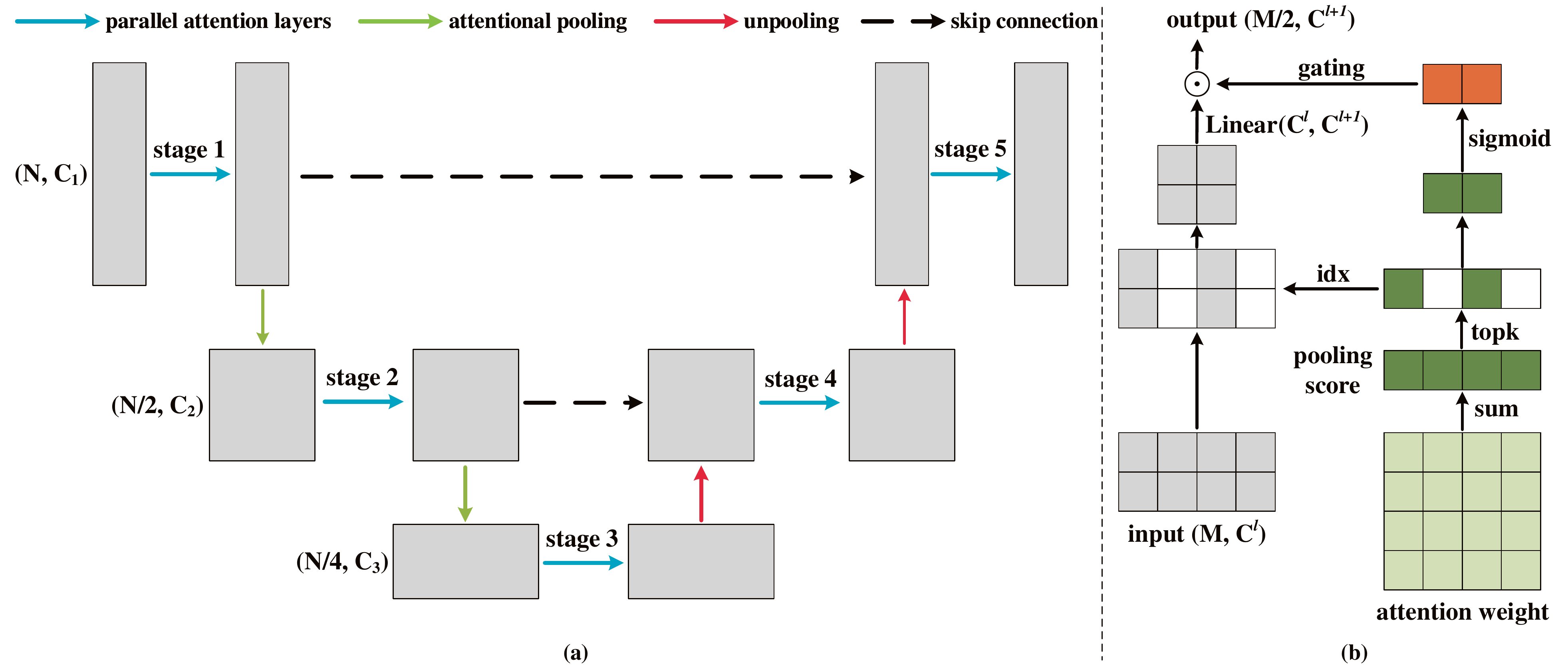} 
    \caption{(a) U-Net architecture. The descriptors are processed in an encoder-decoder fashion.
	After the stages 1 and 2, the descriptors are downsampled with attention pooling to filter out the insignificant descriptors.
	After the stages 4 and 5, the descriptors are upsampled and fused with descriptors in previous stage by skip connections.
	(b) Attentional pooling. Pooling scores are computed from attention weights to identify context points and provide the gating signal through \emph{sigmoid} function.}
    \label{figure4}
\end{figure*}

\subsection{Wave Position Encoder}
\label{Wave position encoder}
For the MLP-based position encoder (MLP-PE), the main drawback is the limited encoding capacity because
the parameters of MLP-PE are less than 1$\%$ of the whole network, yet the position information is important for feature matching. 
Therefore, Wave-PE is designed to dynamically adjust the relationship between descriptor and position by amplitude and phase to obtain better position encoding.

In Wave-PE, position encoding is represented as a wave $\tilde{\boldsymbol{w}}$ with both amplitude $\mathbf{A}$ and phase $\boldsymbol{\theta}$ information,
and the Euler formula is employed to unfold waves into real parts and imaginary parts to process waves efficiently,
\begin{equation}
\begin{split}
	\tilde{\boldsymbol{w}}_{j} & =\mathbf{A}_{j} \odot e^{i\boldsymbol{\theta} _{j} } \\
	 & =\mathbf{A}_{j} \odot \cos \boldsymbol{\theta}_{j}+i\mathbf{A}_{j} \odot \sin \boldsymbol{\theta}_{j} ,j=1,2,...,n.
	\label{equ1} 
\end{split}
\end{equation}
As shown in Figure \ref{figure3}, the amplitude and phase are estimated by two learnable networks via descriptors and position, respectively.
Then a learnable network is applied to fuse the real and imaginary parts into position encoding,
\begin{equation}
\begin{split}
	&\mathbf{A}_{j}= MLP_{A}(\mathbf{d} _{j}), \\
	&\boldsymbol{\theta} _{j}= MLP_{\theta}(\mathbf{p}_{j}), \\
	&\textbf{x}_{j}^{0}= \textbf{d}_{j}+MLP_{F}([\mathbf{A}_{j} \odot \cos \boldsymbol{\theta}_{j},\mathbf{A}_{j} \odot \sin \boldsymbol{\theta}_{j}]).
	\label{equ2} 
\end{split}
\end{equation}

\noindent$[\cdot,\cdot]$ denotes concatenation.
For three learnable networks in equation (2), two-layer MLP is chosen for simplicity.

\subsection{Parallel Attention}
\label{Parallel attention}
As illustrated in the right side of Figure \ref{figure3}, the two sets of descriptors are first linearly projected as $\boldsymbol{Q}, \boldsymbol{K}, \boldsymbol{V}$.
Then self- and cross-attention are computed in a parallel manner. 
In the self-attention module, standard attention computation $softmax(\boldsymbol{Q}\boldsymbol{K}^{T}/\sqrt{d})\boldsymbol{V}$ is employed,
where $\boldsymbol{Q}, \boldsymbol{K}, \boldsymbol{V}$ come from the same input, \emph{i.e.},
($\boldsymbol{Q}_{x}, \boldsymbol{K}_{x}, \boldsymbol{V}_{x}$) or ($\boldsymbol{Q}_{y}, \boldsymbol{K}_{y}, \boldsymbol{V}_{y}$).
In the cross-attention module, the attention weight sharing strategy is proposed to improve model efficiency, which is replacing $\boldsymbol{Q}_{y}\boldsymbol{K}_{x}^{T}$ with $(\boldsymbol{Q}_{x}\boldsymbol{K}_{y}^{T})^{T}$,
so the input of the cross-attention module is ($\boldsymbol{Q}_{x}, \boldsymbol{V}_{x}, \boldsymbol{K}_{y}, \boldsymbol{V}_{y}$).
The impact of weight sharing and attention weight sharing is investigated in the ablation studies.
Finally, self- and cross-attention outputs are fused by a two-layer MLP.
Parallel attention saves redundant parameters and computations while boosting performance through learnable fusion.

Since the parallel attention layer updates two sets of descriptors simultaneously,
it is formally similar to the self-attention layer in mainstream Transformers,
except with two inputs. We can simply stack \emph{L} parallel attention layers to form ParaFormer
and conveniently explore various architectures like U-Net architecture to design model variants.

\subsection{U-Net Architecture}
\label{U_net architecture}
As shown in Figure \ref{figure4} (a), ParaFormer-U is designed for efficiency.
Spatial downsampling is performed first to extract the high-level semantic information,
then upsampling is performed to recover the spatial information, and the low-level and high-level information are fused by skip connections.

As illustrated in Figure \ref{figure4} (b), attentional pooling is proposed for downsampling.
Observing the attention map by column shows that certain points have strong weight with all other points,
indicating that they are important context points in the image.
Suppose the feature in layer \emph{l} is $\textbf{x}^{l}\in\mathbb{R}^{N\times D}$ and the attention weight is $\textbf{A}^{l}\in\mathbb{R}^{N\times N}$.
Our proposed attentional pooling is defined as
\begin{equation}
	\begin{split}
		&\textbf{s}  = sum(\textbf{A}, dim=1), \\
		&\textbf{idx}  = rank(\textbf{s},\textbf{k}), \\
		&\tilde{\textbf{x}}^{l}  = Linear(\textbf{x}^{l}(\textbf{idx}, :)), \\
		&\textbf{g}  = sigmoid(\textbf{s}(\textbf{idx})) \\
		&\textbf{x}^{l+1}  = \tilde{\textbf{x}}^{l} \odot \textbf{g}.
		\label{equ5} 
	\end{split}
\end{equation}
\textbf{k} is the number of points selected for next layer \emph{l}+1, which is set to half the number of points in previous layer \emph{l}.
The sum of each column of the self-attention map is computed as the pooling score $\textbf{s}\in\mathbb{R}^{N}$, which measures the importance of each point.
Then topk points are selected based on the attentional pooling score to filter out insignificant points, and the \emph{Linear} layer is used to adjust the dimension size of the descriptors.
\textbf{s}(\textbf{idx}) extracts values in \textbf{s} with indices \textbf{idx} followed by a \emph{sigmoid} operation to generate gating signal,
and $\odot$ represents the element-wise matrix multiplication.

Following \cite{gu_net}, the unpooling operation is defined as
\begin{equation}
	\begin{split}
		&\tilde{\textbf{x}}^{l}  = Linear(\textbf{x}^{l}), \\
		&\textbf{x}^{l+1}  = distribute(\textbf{0}_{N\times C^{l+1}}, \tilde{\textbf{x}}^{l}, \textbf{idx}),
		\label{equ7} 
	\end{split}
\end{equation}
where $\textbf{x}^{l}\in\mathbb{R}^{k\times C^{l}}$ is the current feature matrix and $\textbf{0}_{N\times C^{l+1}}$ initially empty feature matrix for the next layer.
The \emph{Linear} layer is employed first to adjust the feature matrix dimension.
$\textbf{idx}\in\mathbb{R}^{k}$ is the indices of points selected in the corresponding pooling layer.
Then the current feature matrix is inserted into the corresponding row of the empty feature matrix according to \textbf{idx}, while the other rows remain zero.
In other words, the unselected features in the pooling layer are represented by zero vectors to perform upsampling.

\subsection{Implementation Details}
\label{Implementation Details}
The homography model is pretrained on the $\mathcal{R}$1M dataset \cite{r1m},
and then the model is finetuned on the MegaDepth dataset \cite{8578316} for outdoor pose estimation and image matching tasks.
On the $\mathcal{R}$1M dataset, we employ the AdamW \cite{adam} optimizer for 10 epochs using the cosine decay learning rate scheduler and 1 epoch of linear warm-up.
A batch size of 8 and an initial learning rate of 0.0001 are used.
On the MegaDepth dataset, we use the same AdamW optimizer for 50 epochs using the same learning rate scheduler and linear warm-up.
A batch size of 2 and a lower initial learning rate of 0.00001 are used.
For training on $\mathcal{R}$1M/MegaDepth dataset, we resize the images to 640$\times$480/960$\times$720 pixels and detect 512/1024 keypoints, respectively.
When the detected keypoints are not enough, random keypoints are added for efficient batching.
All models are trained on a single NVIDIA 3070Ti GPU.
For ParaFormer, we stack $L=9$ parallel attention layers, and all intermediate features have the same dimension $C = 256$.
For ParaFormer-U, the depth of each stage is $\left \{ 2, 1, 2, 1, 2 \right \}$, resulting in a total of $L=8$ parallel attention layers, and the intermediate feature dimension of each stage is $\left \{ 256, 384, 128, 384, 256 \right \}$.
More details are provided in the supplementary material.

\begin{table}[t]
	\large
	\centering
	\resizebox{.92\columnwidth}{!}{
	\begin{tabular}{@{}lcccc@{}}
	\toprule
		Matcher                                       & AUC                                          & Precision                                       & Recall                                   & F1-score           \\ \midrule
		NN                                            & 39.47                                         & 21.7                                            & 65.4                                     & 32.59                     \\
		NN + mutual                                   & 42.45                                         & 43.8                                            & 56.5                                     & 49.35    \\
		NN + PointCN                                  & 43.02                                         & 76.2                                            & 64.2                                     & 69.69    \\
		NN + OANet                                    & 44.55                                         & 82.8                                            & 64.7                                     & 72.64    \\
		SuperGlue                                     & 52.65        								    & 90.9    									      & 98.88 								   & 94.72     \\
		\textbf{ParaFormer-U}    						& 53.16        		    						& 90.93    			      						  & 99.01 	 						           & 94.80						  \\
		\textbf{ParaFormer}         						& \textbf{54.91}								& \textbf{94.55}		  						  & \textbf{99.10}     		 				  & \textbf{96.77} \\ \bottomrule  
	\end{tabular}
	}
	\caption{ 
	Homography estimation on $\mathcal{R}$1M. AUC @10 pixels is reported. The best method is highlighted in bold.}
	\label{table1}
\end{table}

\section{Experiments}
\label{Experiments}

\subsection{Homography Estimation}
\label{Homography Estimation}

\noindent\textbf{Dataset.}
We split $\mathcal{R}$1M dataset \cite{r1m}, which contains over a million images of Oxford and Paris, into training, validation, and testing sets.
To perform self-supervised training, random ground-truth homographies are generated to get image pairs.

\noindent\textbf{Baselines.}
SuperPoint \cite{superpoint} is applied as the unified descriptor to generate the input for the matcher.
ParaFormer and ParaFormer-U are compared with attention-based matcher SuperGlue \cite{9157489} and NN matcher with learning-based outlier rejection methods \cite{8578380,9310246}.
The results of SuperGlue are from our own implementation.

\noindent\textbf{Metrics.}
Precision and recall are computed based on ground truth matches.
The area under the cumulative error curve (AUC) up to a value of 10 pixels is reported, where the reprojection error is computed with the estimated homography.

\noindent\textbf{Results.} 
As shown in Table \ref{table1}, ParaFormer outperforms all outlier rejection methods and attention-based matcher on homography estimation. 
It can be seen that the attention-based approaches have a remarkable superiority due to the global receptive field of attention. 
Compared with the attention-based approach SuperGlue, ParaFormer further boosts the performance by integrating self- and cross-attention with parallel attention layers, bringing a $+2.05\%$ improvement on the F1-score over SuperGlue.
The visualization of matches can be found in Figure \ref{figure6}.
Moreover, compared to SuperGlue, our efficient U-Net variant has only $49\%$ FLOPs, yet achieves better performance.

\begin{table}[t]
	\centering
	\LARGE
	\resizebox{.99\columnwidth}{!}{
	\begin{tabular}{@{}lcccccc@{}}
	\toprule
	\multirow{2.5}{*}{Matcher} & \multicolumn{3}{c}{Exact AUC} & \multicolumn{3}{c}{Approx. AUC} \\ \cmidrule(l){2-7} 
							 & @5$^{\circ}$       & @10$^{\circ}$      & @20$^{\circ}$     & @5$^{\circ}$        & @10$^{\circ}$      & @20$^{\circ}$       \\ \midrule
	NN + mutual              & 16.94    & 30.39    & 45.72   & 35.00     & 43.12    & 54.25    \\
	NN + OANet               & 26.82    & 45.04    & 62.17   & 50.94     & 61.41    & 71.77    \\
	SuperGlue                & 28.45    & 48.6     & 67.19   & 55.67     & 66.83    & 74.58    \\
	\textbf{ParaFormer-U}       & 29.40    & 49.76    & 68.29   & 56.47     & 67.66    & 75.67    \\
	\textbf{ParaFormer}            & \textbf{31.73}    & \textbf{52.28}    & \textbf{70.43}   & \textbf{60.05}     & \textbf{70.72}    & \textbf{78.13}    \\ \bottomrule
	\end{tabular}%
	}
	\caption{
	Pose estimation on YFCC100M. ParaFormer and ParaFormer-U lead other methods at all thresholds.}
	\label{table2}
\end{table}

\begin{figure*}[t]
    \centering
    \includegraphics[width=0.99\textwidth]{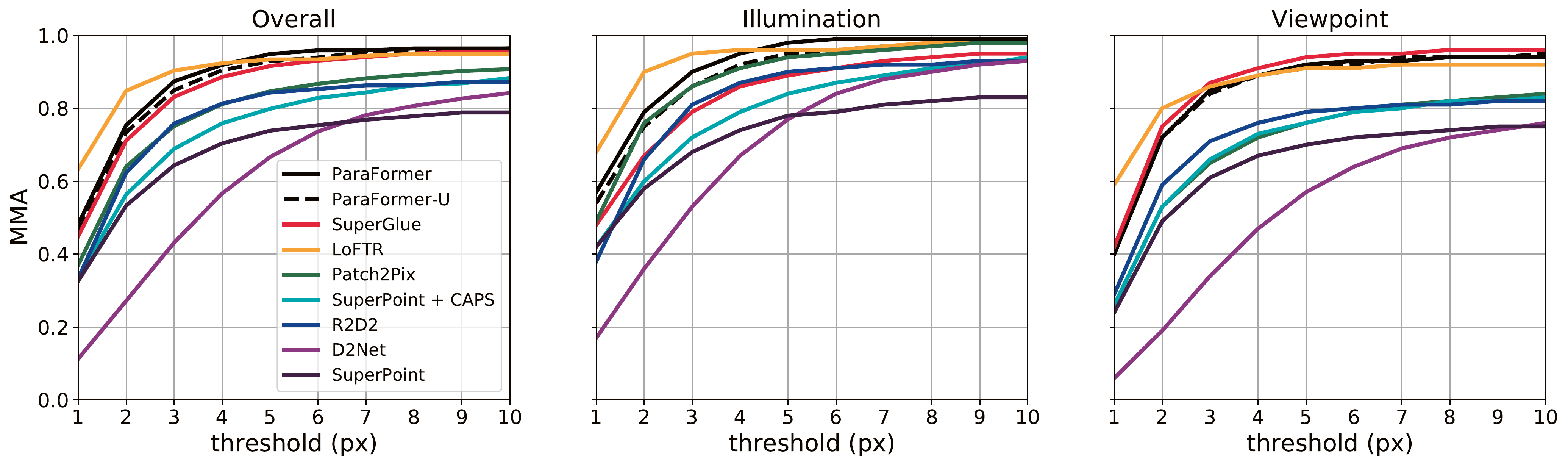} 
    \caption{
	Image matching on HPatches. The mean matching accuracy (MMA) at thresholds from 1 to 10 pixels are reported.}
    \label{figure5}
\end{figure*}

\subsection{Outdoor Pose Estimation}
\label{Outdoor pose estimation}

\noindent\textbf{Dataset.}
ParaFormer is trained on the MegaDepth dataset \cite{8578316} and evaluated on the YFCC100M dataset \cite{2016YFCC100M}.
For training, 200 pairs of images in each scene are randomly sampled for each epoch. For evaluation,
the YFCC100M image pairs and ground truth poses provided by SuperGlue are used. 

\noindent\textbf{Baselines.}
SuperPoint is applied as the descriptor and combined with baseline matchers, which contain attention-based matcher SuperGlue and NN matcher with outlier rejection methods \cite{2004Distinctive,9310246}.
The results of SuperGlue are from our own implementation.

\noindent\textbf{Metrics.}
The AUC of the pose error at thresholds ($5^\circ$, $10^\circ$, $20^\circ$) are reported. 
Evaluation is performed with both approximate AUC \cite{9310246} and exact AUC \cite{9157489} for a fair comparison.

\noindent\textbf{Results.} 
As shown in Table \ref{table2}, ParaFormer achieves the best performance at all thresholds, demonstrating the robustness of our models.
With wave position encoder and parallel attention architecture, ParaFormer can bring $(+3.28\%, +4.2\%, +3.24\%)$ improvement on exact AUC
and $(+4.38\%, +3.89\%, +3.55\%)$ improvement on approximate AUC at three thresholds of $(5^{\circ}, 10^{\circ}, 20^{\circ})$, respectively.
In outdoor scenes with a large number of keypoints, ParaFormer-U can effectively alleviate the computational complexity problem by downsampling,
while still maintaining state-of-the-art performance by attentional pooling.

\subsection{Image Matching}
\label{Image Matching}

\noindent\textbf{Dataset.}
We follow the evaluation protocol as in D2-Net \cite{d2net} and evaluate our methods on 108 HPatches \cite{hpatches} sequences, which contain 52 sequences with illumination changes and 56 sequences with viewpoint changes.

\noindent\textbf{Baselines.}
Baseline methods include learning-based descriptors R2D2, D2Net and SuperPoint \cite{r2d2, d2net,superpoint} and advanced matchers LoFTR, Patch2Pix, SuperGlue and CAPS \cite{loftr,patch2pix,9157489,caps}.
The results of SuperGlue are from our own implementation.

\noindent\textbf{Metrics.}
A match is considered correct if the reprojection error is below the matching threshold, where the reprojection error is computed from the homographies provided by the dataset.
The matching threshold is varied from 1 to 10 to plot the mean matching accuracy (MMA), which is the average percentage of correct matches for each image.

\begin{table}[t]
	\Huge
	\centering
	\resizebox{.90\columnwidth}{!}{
	\begin{tabular}{@{}ccccccc@{}}
	\toprule
	SA & \textbf{PA} & MLP-PE & \textbf{Wave-PE} & Precision & Recall & F1-score \\ \midrule
	\checkmark  &    & \checkmark      &         & 86.68     & 96.56  & 91.35    \\
	& \checkmark  & \checkmark      &         & 86.79     & 98.23  & 92.16    \\
	& \checkmark  &        & \checkmark       & \textbf{87.69}     & \textbf{98.73}  & \textbf{92.88}    \\ \bottomrule
	\end{tabular}%
	}
	\caption{
	Ablation study on main designs.}
	\label{table3}
\end{table}


\begin{table}[t]
	\Huge
	\centering
	\resizebox{0.85\linewidth}{!}{%
	\begin{tabular}{@{}ccccc@{}}
	\toprule
	FFN & QKV proj & Head Merging & \#Params (M)    			    & F1-score       \\ \midrule
		&          &              & 1.70          					& 88.79          \\ 
		\checkmark   &          &  & \textbf{1.18}  				        & 80.86          \\
		& \checkmark        &              & 1.51   			& 90.15          \\
		&          & \checkmark            & 1.51   			    & \textbf{90.16} \\
		& \checkmark        & \checkmark   & 1.31 					& 90.02          \\ \bottomrule
	\end{tabular}%
	}
	\caption{
	Ablation study on weight sharing.
	}
	\label{table4}
\end{table}

\noindent\textbf{Results.} 
As shown in Figure \ref{figure5}, ParaFormer achieves the best overall performance at matching thresholds of 5 or more pixels.
The results indicate that detector-based methods such as SuperGlue and our methods are better at handling scenarios with large viewpoint changes,
while the detector-free methods such as LoFTR are better suited to address illumination changes.
But ParaFormer still outperforms LoFTR in illumination change experiments, benefiting from the superior modeling capability of parallel attention.
With ParaFormer and ParaFormer-U, the overall performance of SuperPoint grows from the last to the first and the second place, demonstrating the effectiveness of our matchers.

\begin{figure*}[t]
    \centering
    \includegraphics[width=0.95\textwidth]{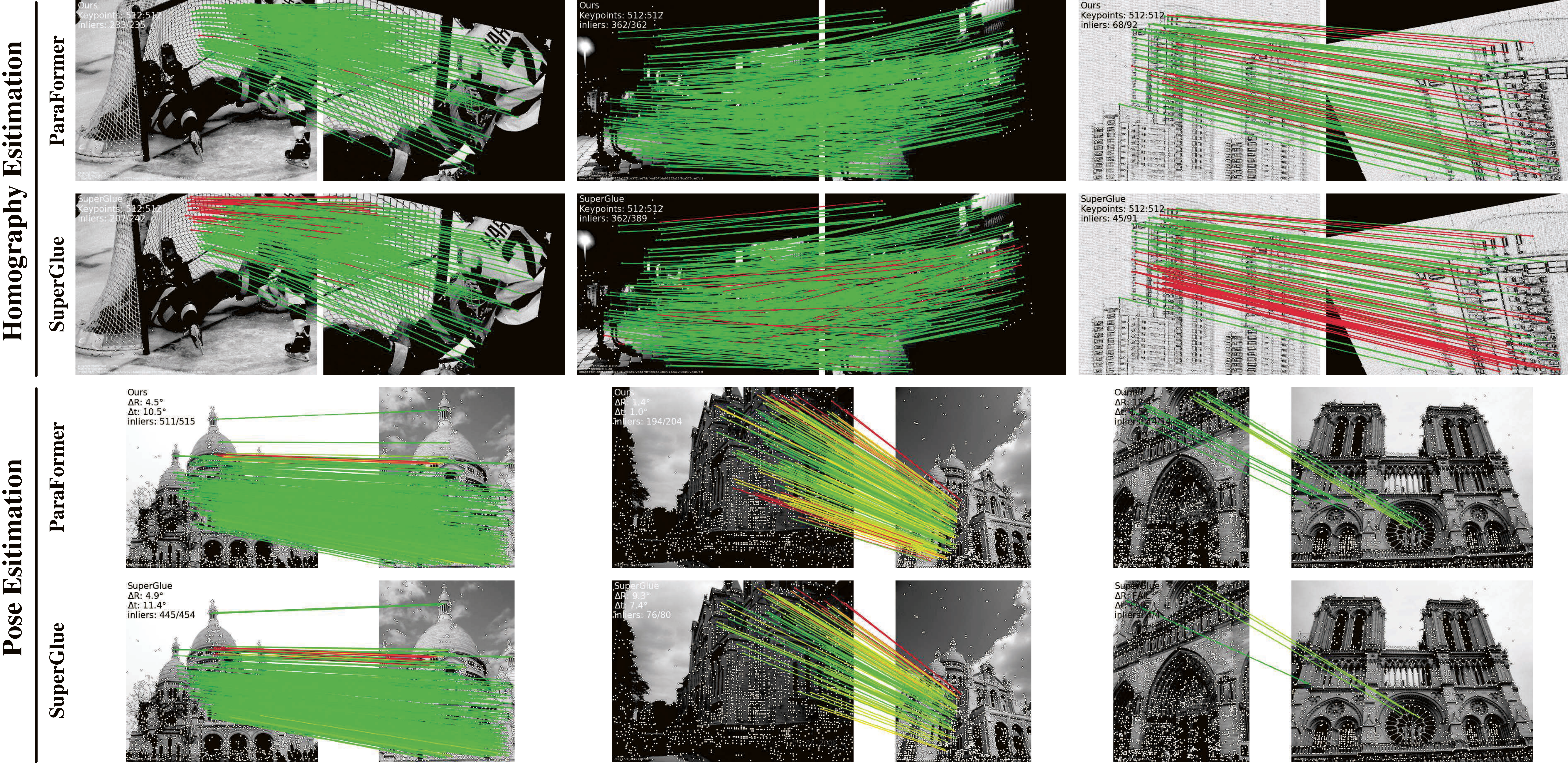} 
    \caption{Qualitative results of homography estimation and pose estimation experiments.}
    \label{figure6}
\end{figure*}

\begin{figure}[t]
	\centering
	\includegraphics[width=0.99\columnwidth]{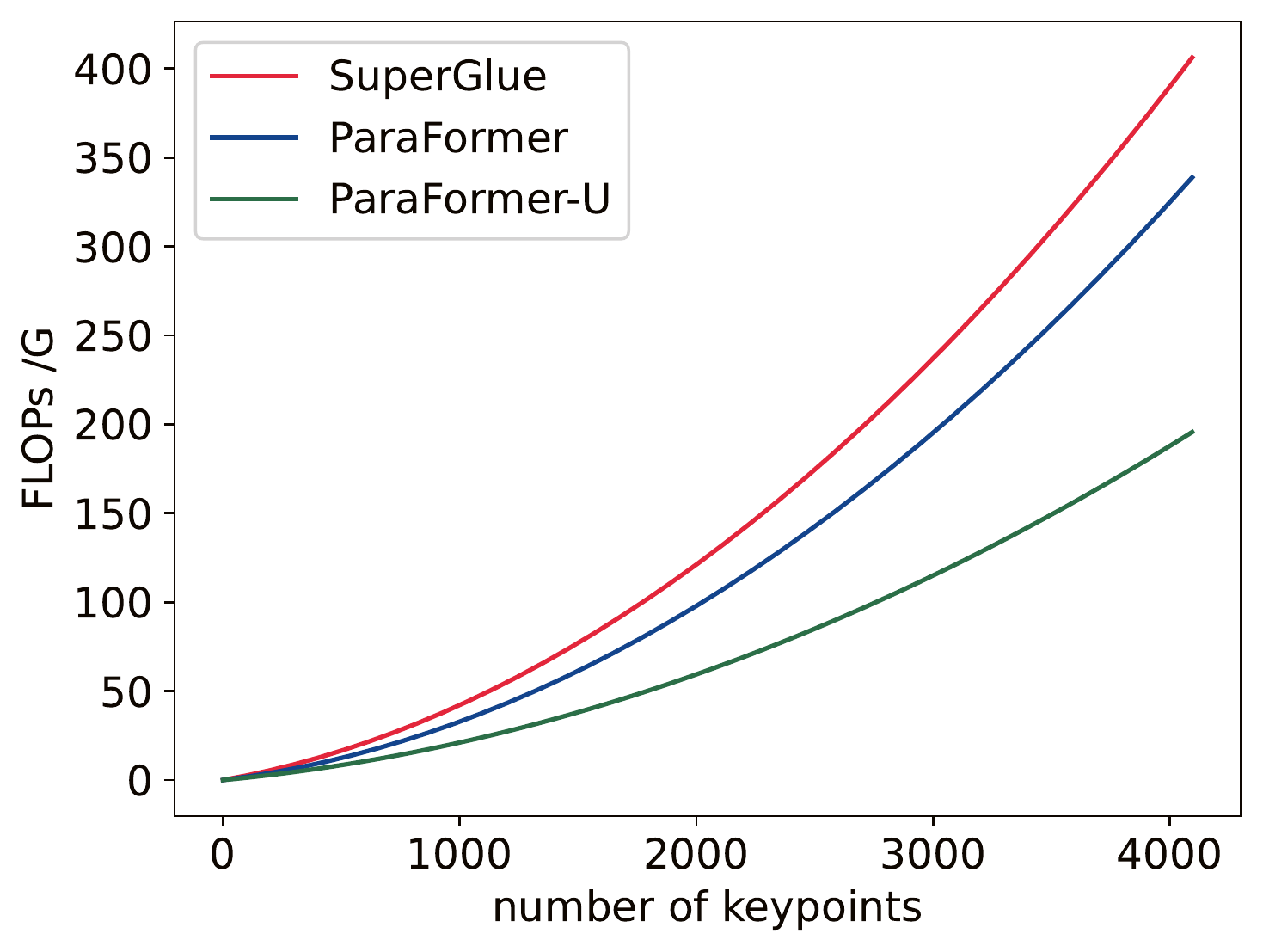}
	\caption{
	Comparison between FLOPs of models.}
	\label{figure7}
\end{figure}

\begin{table}[t]
	\centering
	\resizebox{0.82\linewidth}{!}{%
	\begin{tabular}{@{}ccccc@{}}
	\toprule
	attention weight sharing & FLOPs (G)   & F1-score \\ \midrule
							 & 108.72        & 89.98          \\
	\checkmark               & \textbf{99.05}     & \textbf{90.02}    \\ \bottomrule
	\end{tabular}%
	}
	\caption{
	Ablation study on attention weight sharing.}
	\label{table5}
\end{table}

\begin{table}[t]
	\Huge
	\centering
	\resizebox{0.87\linewidth}{!}{%
	\begin{tabular}{@{}cccc@{}}
	\toprule
	random pooling & gPool & \textbf{attentional pooling}   & F1-score \\ \midrule
	\checkmark           &       &                                  & 90.87          \\
				& \checkmark     &                         & 90.95    \\
				&       & \checkmark                       & \textbf{91.23}     \\ \bottomrule
	\end{tabular}
	}
	\caption{
	Ablation study on pooling.}
	\label{table6}
\end{table}


\subsection{ParaFormer Structural Study}
\label{ParaFormer Structural Study}
A complete ablation study is conducted on the $\mathcal{R}$1M dataset for further understanding of our designs.
The FLOPs and runtimes between our methods and SuperGlue are also compared to demonstrate the high efficiency of our methods.

\noindent{\textbf{Main Designs.}}
We did ablation experiments on parallel attention architecture and Wave-PE.
As can be seen from Table \ref{table3}, when both use MLP-PE, parallel attention leads serial attention on both precision and recall, resulting in a $+0.81\%$ improvement on F1-score.
When parallel attention is combined with Wave-PE, the performance can be further boosted by $+0.72\%$ over MLP-PE, indicating that Wave-PE provides stronger position information to guide matching.

\noindent{\textbf{Weight Sharing.}}
As shown in Table \ref{table4}, we conduct ablation experiments on weight sharing strategies and find that the performance of the network improved when self-attention and cross-attention
share the $\boldsymbol{Q}, \boldsymbol{K}, \boldsymbol{V}$ projection weights and the multi-head merging weights, while it also helps to reduce the model parameters.
This occurs because self- and cross-attention are essentially indistinguishable except for the inputs,
and the shared weights align the $\boldsymbol{Q}, \boldsymbol{K}, \boldsymbol{V}$ projections of both inputs so that self- and cross-attention can be performed in the same vector space, which makes a uniform standard for the cosine similarity of both.

\noindent{\textbf{Attention Weight Sharing.}}
We find that computing cross-attention twice is redundant for attention-based methods,
because of the high correlation between $\boldsymbol{Q}_{y}\boldsymbol{K}_{x}^{T}$ and $\boldsymbol{Q}_{x}\boldsymbol{K}_{y}^{T}$.
So attention weight sharing is proposed for efficiency, \emph{i.e.},
replacing $\boldsymbol{Q}_{y}\boldsymbol{K}_{x}^{T}$ with $(\boldsymbol{Q}_{x}\boldsymbol{K}_{y}^{T})^{T}$.
As shown in Table \ref{table5}, attention weight sharing can reduce FLOPs without performance loss,
which makes a significant difference in scenarios with a large number of keypoints.

\noindent{\textbf{Attentional Pooling.}}
As shown in Table \ref{table6}, compared to gPool \cite{gu_net} which gets pooling scores by linear projection,
our proposed attentional pooling achieves better performance and saves the parameters of linear projection by identifying important context points by attention weights.
As expected, the strategy of computing pooling scores by features or attention weights is superior to random pooling.

\noindent{\textbf{Efficiency Analysis.}}
Benefiting from the above designs, our model is remarkable in efficiency beyond just achieving state-of-the-art performance.
As shown in Table \ref{table7}, when matching 2048 descriptors, ParaFormer reduces FLOPs by $17.8\%$ compared to SuperGlue with better performance.
ParaFormer-U further improves efficiency with FLOPs of only $49\%$ of SuperGlue, while it still outperforms SuperGlue due to the advantage of parallel attention and Wave-PE.
As shown in Figure \ref{figure7}, the attention weight sharing strategy in ParaFormer alleviates the squared complexity of the attention mechanism,
and the U-Net architecture further significantly reduces computational cost through downsampling.

\begin{table}[t]
	\centering
	\resizebox{0.9\linewidth}{!}{%
	\begin{tabular}{@{}lccc@{}}
		\toprule
		Methods       								& F1-score          & FLOPs (G)          & Runtime (ms)        \\ \midrule
		SuperGlue     										& 90.68     		   & 125.85            & 26.99          \\
		\textbf{ParaFormer-U}  								& 90.72          	   & \textbf{61.67}    & \textbf{20.23} \\
		\textbf{ParaFormer} 				& \textbf{94.92}       & 104.22 		   & 24.99     \\ \bottomrule
	\end{tabular}%
	}
	\caption{
	Efficiency analysis @2048 keypoints.}
	\label{table7}
\end{table}


\section{Conclusion}
In this paper, we propose a novel attention-based network named ParaFormer to handle feature matching tasks efficiently.
As a preprocessing module, the proposed Wave-PE dynamically fuses features and positions in amplitude and phase manner.
In contrast to employing serial attention that intuitively mimics human behavior,
we propose a parallel attention architecture that not only integrates self-attention and cross-attention in a learnable way
but also saves redundant parameters and computations through weight sharing and attention weight sharing strategies.
To further improve efficiency,
the ParaFormer-U is designed with U-Net architecture, which reduces FLOPs by downsampling and minimizes the performance loss by the proposed attentional pooling.
Experiments show that ParaFormer and ParaFormer-U deliver state-of-the-art performance with remarkable efficiency,
enabling a broader application scenario for attention-based feature matching networks.

\section{Acknowledgments}
This work was jointly supported by the National Natural Science Foundation of China under grants 62001110, 62201142 and 62171232, the Natural Science Foundation of Jiangsu Province under grant BK20200353, and the China Postdoctoral Science Foundation under grant 2020M681684.

\bibliography{1476.LuX.bib}
\end{document}